\documentclass[runningheads]{llncs}
\usepackage{graphicx}
\usepackage{comment}
\usepackage{amsmath,amssymb} 
\usepackage{color}

\usepackage{stfloats}
\usepackage{xspace}
\usepackage{adjustbox}
\usepackage{array}
\usepackage{hyperref} 
\usepackage{cleveref}
\usepackage{amsfonts}
\usepackage{booktabs}
\usepackage{siunitx}
\usepackage{multirow}
\usepackage{colortbl}
\usepackage[table, svgnames, dvipsnames]{xcolor}

\usepackage[width=122mm,left=12mm,paperwidth=146mm,height=193mm,top=12mm,paperheight=217mm]{geometry}

\begin{document}

\pagestyle{headings}
\mainmatter
\def\ECCVSubNumber{100}  

\title{KPRNet: Improving projection-based LiDAR semantic segmentation} 



\titlerunning{KPRNet}
%
\author{Deyvid Kochanov \and
Fatemeh Karimi Nejadasl \and
Olaf Booij}
\authorrunning{Kochanov et al.}
%
\institute{TomTom, Amsterdam, Netherlands
\email{\{deyvid.kochanov,fatemeh.kariminejadasl,olaf.booij\}@tomtom.com}}

\maketitle

\begin{abstract}

Semantic segmentation is an important component in the perception systems of autonomous vehicles. In this work, we adopt recent advances in both image and point cloud segmentation to achieve a better accuracy in the task of segmenting LiDAR scans. KPRNet improves the convolutional neural network architecture of 2D projection methods and utilizes KPConv to replace the commonly used post-processing techniques with a learnable point-wise component which allows us to obtain more accurate 3D labels. With these improvements our model outperforms the current best method on the SemanticKITTI benchmark, reaching an mIoU of 63.1.


\end{abstract}

\section{Introduction}
Semantic segmentation plays an important role in the autonomous driving perception stack. This started mainly with the segmentation of camera-based images, but since the release of the SemanticKITTI dataset~\cite{SemanticKITTI} significant progress has been made in the segmentation of LiDAR measurements.

Current LiDAR based semantic segmentation can roughly be categorized in two approaches, which reach a comparable performance. The first approach, uses purely point-wise methods acting directly on the 3D point cloud~\cite{KPConv,randla}. The second approach builds on top of the well developed field of image segmentation, which focuses on CNN architectures for segmenting RGB images~\cite{zhao2017pyramid,chen2017rethinking,yuan2019object,sun2019high}. To this end, individual LiDAR sweeps are projected to 2D range images, which then serve as input to custom CNNs~\cite{rangenet,SalsaNext,squeezesegv3}. The resulting 2D predictions are then post-processed with non-learned CRFs or KNN-based voting to recover more accurate labels for each 3D point.

In this work, we combine the best of both these approaches. The main contribution is twofold. First, we propose an improved CNN architecture for 2D projected LiDAR sweeps. Next, we replace the post-processing step with a learnable module based on KPConv~\cite{KPConv}.

On the SemanticKITTI benchmark the resulting model outperforms the current best 2D segmentation method SalsaNext~\cite{SalsaNext} by 3.6 mIoU points and the best point-wise method which relies on KPConv by 4.3 mIoU points.\footnote{20-July-2020 submitted to CodaLab. The name is not\_bad\_jpg.}

\section{Method}
The proposed method combines a 2D semantic segmentation network (Section~\ref{2D}), and a 3D point-wise layer. The convolutional network gets as input a LiDAR scan projected to a range image. The resulting 2D CNN features are projected back to their respective 3D points and passed to a 3D point-wise module (Section~\ref{3D}), which predicts the final labels.

\begin{figure}
\centering
\includegraphics[height=3.5cm]{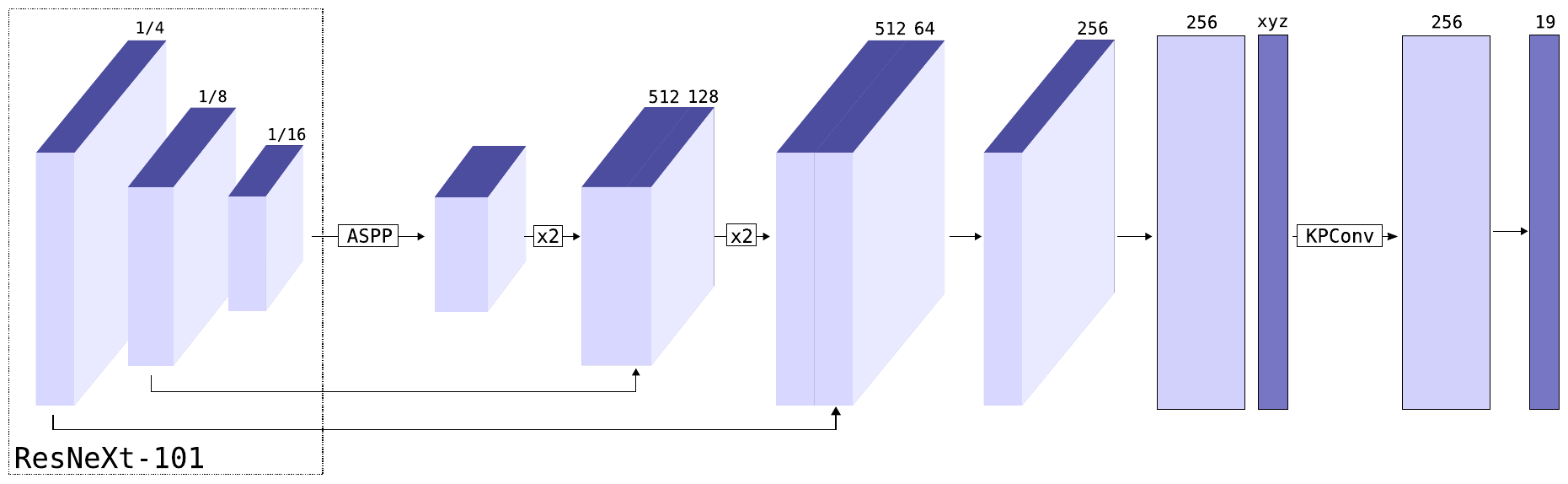}
\caption{The KPRNet architecture: ResNeXt features with stride 16 are fed into an ASPP module~\cite{chen2017rethinking} and combined with the outputs of the second and first ResNeXt blocks, which have strides of 8 and 4. The result is passed through a KPConv layer followed by BatchNorm, ReLu and a final classifier.}
\label{fig:architecture}
\end{figure}

\subsection{2D semantic segmentation}\label{2D}

In our model, we use a ResNeXt-101 encoder~\cite{mahajan2018exploring} and a decoder similar to Panoptic-DeepLab~\cite{cheng2020panoptic} (see Figure~\ref{fig:architecture}). The CNN part of our final architecture is pre-trained on Cityscapes~\cite{Cityscapes}. When transferring the model to the LiDAR segmentation task, we discard one of the filter planes in the first layer  because the input for the task requires only 2 channels - inverse depth and reflectivity.

\subsection{3D semantic segmentation}\label{3D}
To generate 2D range images most methods~\cite{rangenet,SalsaNext,squeezesegv3} perform a spherical projection on the point cloud. 
In~\cite{triess2020scan} an alternative method was proposed which unfolds the scans according to the order in which they are captured by the LiDAR sensor. This results in smoother projections and we use it in all our experiments. 

Despite this, the remaining discretization artifacts and overly-smooth 2D labels generated by the CNN result in misprediction when back-projecting to the 3D point cloud. RangeNet++~\cite{rangenet} and other 2D segmentation methods perform a KNN or CRF post-processing step to account for this but finding a good balance between over- and under-smoothing the 3D labels with these methods can be difficult.

We replace this post-processing step by inserting a single KPConv~\cite{KPConv} layer before the final classification (see Figure~\ref{fig:architecture}). KPConv is a point-convolution operator which can learn to correct the misclassifications by taking into account the 2D features of each point and its surrounding 3D context. The required modification to the CNN architecture is minimal and the pipeline from a 2D range image to 3D point labels is end-to-end learnable.

\section{Experiments}
\subsection{Setting}
We train and evaluate our model on SemanticKITTI which contains 21 sequences. Sequences 11-21 are reserved as a test split and only available as the official benchmark. Sequence 8 is used for validation and the rest is used for training. \footnote{Code is available at: \url{https://github.com/DeyvidKochanov-TomTom/kprnet}} 

All the models are trained with SGD with momentum of 0.9 and 1e-4 weight decay for 120 epochs. We use a cosine schedule~\cite{cosine} with warm-up in the first 1000 iterations. For the KNN experiments we used a batch size of 32 and a base learning rate of 0.025. The KPConv models are trained with a batch size of 24 and the base learning rate is set to 0.01875. During training random crops of width 1025 are sampled from the range images and random horizontal flipping is applied. All models are trained on 8 Tesla V100-SXM2-16GB GPUs.

\subsection{Quantitative evaluation}

\begin{table*}\centering
\scalebox{0.68}{
\begin{tabular}{l|ccccccccccccccccccc|c}
\toprule
Approach  
&\rotatebox{90}{car}		&\rotatebox{90}{bicycle}		&\rotatebox{90}{motorcycle}		&\rotatebox{90}{truck}			&\rotatebox{90}{other-vehicle}
&\rotatebox{90}{person} 	&\rotatebox{90}{bicyclist}		&\rotatebox{90}{motorcyclist} 	&\rotatebox{90}{road}			&\rotatebox{90}{parking} 	
&\rotatebox{90}{sidewalk}   &\rotatebox{90}{other-ground} 	&\rotatebox{90}{building} 		&\rotatebox{90}{fence} 			&\rotatebox{90}{vegetation} 
&\rotatebox{90}{trunk} 		&\rotatebox{90}{terrain} 		&\rotatebox{90}{pole} 			&\rotatebox{90}{traffic-sign}	&\rotatebox{90}{\textbf{mean-IoU}}    
\\ 
\hline   

SalsaNext~\cite{SalsaNext} &
91.9 & 48.3 & 38.6 & 38.9 & 31.9 & 60.2 & 59.0 & 19.4 & 91.7 & 63.7 & 75.8 & 29.1 & 90.2 & 64.2 & 81.8 & 63.6 & 66.5 & 54.3 & 62.1 & 59.5\\
KPConv~\cite{KPConv} &
\textbf{96.0} & 30.2 & 42.5 & 33.4 & \textbf{44.3} & 61.5 & 61.6 & 11.8 & 88.8 & 61.3 & 72.7 & \textbf{31.6} & 90.5 & 64.2  &84.8 & 69.2 & 69.1 & 56.4 & 47.4 & 58.8\\
SqueezeSegV3~\cite{squeezesegv3} &   
92.5 & 38.7 & 36.5 & 29.6 & 33.0 & 45.6 & 46.2 & \textbf{20.1} & 91.7 & 63.4  & 74.8 & 26.4 & 89.0 & 59.4 & 82.0 & 58.7 & 65.4 & 49.6 & 58.9 & 55.9 \\
RandLa-Net~\cite{randla} & 
94.2 & 26.0 &  25.8 & \textbf{40.1} & 38.9 & 49.2 & 48.2 &7.2 & 90.7 & 60.3 & 73.7 & 20.4 & 86.9 & 56.3 & 81.4 & 61.3 & 66.8 & 49.2 & 47.7 & 53.9 \\ 
RangeNet++~\cite{rangenet} &
91.4 & 25.7 & 34.4 & 25.7 & 23.0 & 38.3 & 38.8 & 4.8 & 91.8 & 65.0 & 75.2 & 27.8 & 87.4 & 58.6 & 80.5 & 55.1 & 64.6 & 47.9 & 55.9 & 52.2 \\
%

\hline

KPRNet [Ours] &
95.5 & \textbf{54.1} & \textbf{47.9} & 23.6 & 42.6 & \textbf{65.9} & \textbf{65.0} & 16.5 & \textbf{93.2}  & \textbf{73.9} & \textbf{80.6} & 30.2 & \textbf{91.7} & \textbf{68.4}  & \textbf{85.7} & \textbf{69.8} & \textbf{71.2}  & \textbf{58.7} & \textbf{64.1} & \textbf{63.1}  \\

\bottomrule
\end{tabular}
}
\caption{Quantitative comparison on SemanticKITTI test benchmark.} 
\label{tab:quanresults}
\end{table*}

First we report results on the validation set. We train our CNN architecture and use the KNN post-processing~\cite{rangenet} as a baseline.

The input scans have relatively low vertical resolution of 64x2048. To account for this RangeNet++ uses only horizontal strides in their down-sampling layers. Instead of that, we simply upscale the input range image to 145x2049 using nearest neighbour interpolation. This baseline model achieves 57.7 mIoU. In our second experiment we further upsample the image to 289x4097. This model achieves 61.7 mIoU on the validation set. Next we replace the KNN post-processing with a KPConv layer which results in 64.1 mIoU. 

We train this final model on the combined training and validation subsets and submit to the official SemanticKITTI benchmark. The evaluation results and comparison to state-of-the-art models are show in Table~\ref{tab:quanresults}.

%
%
\bibliographystyle{splncs04}
\bibliography{KPRNet}
\end{document}